\documentclass{article}
\usepackage{ijcai18}

\usepackage{times}
\usepackage{xcolor}
\usepackage{soul}
\usepackage[utf8]{inputenc}
\usepackage[small]{caption}
\usepackage{graphicx}
\usepackage{amsmath}
\usepackage{bm}
\usepackage{multirow}
\usepackage{amsfonts,amssymb}
\usepackage[colorlinks,linkcolor=red,anchorcolor=blue,citecolor=green]{hyperref}
\title{
Co-occurrence Feature Learning from Skeleton Data for Action Recognition and Detection with Hierarchical Aggregation
}

\author{
Chao Li,
Qiaoyong Zhong,
Di Xie,
Shiliang Pu
\\
Hikvision Research Institute \\
\{lichao15, zhongqiaoyong, xiedi, pushiliang\}@hikvision.com
}
\begin{document}

\maketitle

\begin{abstract}
  Skeleton-based human action recognition has recently drawn increasing attentions with the availability of large-scale skeleton datasets. The most crucial factors for this task lie in two aspects: the intra-frame representation for joint co-occurrences and the inter-frame representation for skeletons' temporal evolutions. In this paper we propose an end-to-end convolutional co-occurrence feature learning framework. The co-occurrence features are learned with a hierarchical methodology, in which different levels of contextual information are aggregated gradually. Firstly point-level information of each joint is encoded independently. Then they are assembled into semantic representation in both spatial and temporal domains. Specifically, we introduce a \emph{global} spatial aggregation scheme, which is able to learn superior joint co-occurrence features over local aggregation. Besides, raw skeleton coordinates as well as their temporal difference are integrated with a two-stream paradigm. Experiments show that our approach consistently outperforms other state-of-the-arts on action recognition and detection benchmarks like NTU RGB+D, SBU Kinect Interaction and PKU-MMD.
\end{abstract}

\section{Introduction}

Analysis of human behavior such as action recognition and detection is one of the fundamental and challenging tasks in computer vision. It has a wide range of applications such as intelligent surveillance system, human-computer interaction, game-control and robotics. Articulated human pose, being also referred to as skeleton, provides a very good representation for describing human actions. On one hand, skeleton data are inherently robust against background noise and provide abstract information and high-level features of human action. On the other hand, compared with RGB data, skeleton data are extremely small in size, which makes it possible to design lightweight and hardware friendly models.

\begin{figure}[htbp]
  \centering
  \includegraphics[width=\linewidth]{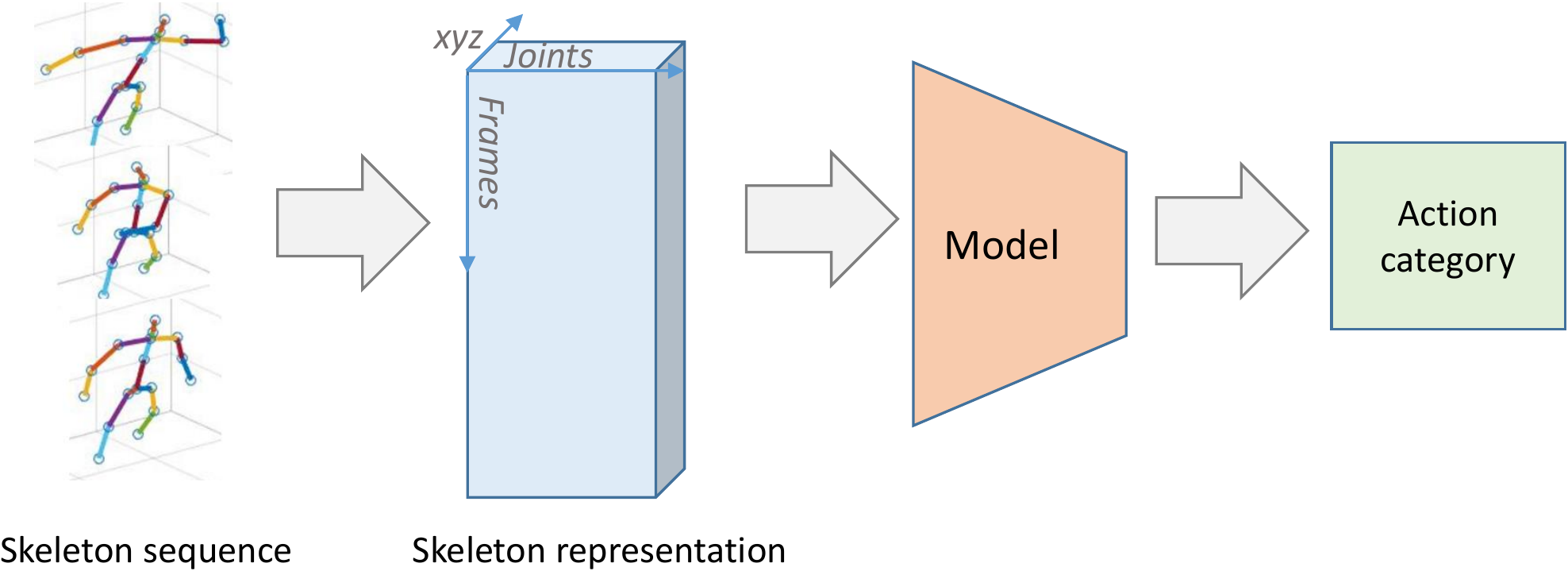}
  \caption{Workflow for skeleton-based human action recognition.}
  \label{fig:workflow}
\end{figure}

In this paper, we focus on the problem of skeleton-based human action recognition and detection (Figure~\ref{fig:workflow}). The interactions and combinations of skeleton joints play a key role to characterize an action. Many of the early works attempted to design and extract co-occurrence features from skeleton sequences, such as pairwise relative position of each joint~\cite{Wang2014Learning}, spatial orientation of pairwise joints~\cite{Jin2012Essential}, and the statistics-based features like Cov3DJ~\cite{Hussein2013Human} and HOJ3D~\cite{hog3D}. On the other hand, the Recurrent Neural Networks (RNNs) with Long-Short Term Memory (LSTM) neurons are used to model the time series of skeleton prevalently~\cite{NTURGBD,song2016end,trust_gate}. Although the LSTM networks were designed to model long-term temporal dependency, it is difficult for them to learn high-level features from skeletons directly since the temporal modeling is done on the raw input space~\cite{Sainath2015}. While fully connected layers are able to learn co-occurrence features for their ability of aggregating global information from all input neurons. In~\cite{co_occurrence}, an end-to-end fully connected deep LSTM network was proposed to learn co-occurrence features from skeleton data.

CNN models are equipped with excellent ability to extract high-level information, and they have been used to learn spatial-temporal features from skeletons~\cite{Du2016Skeleton,Ke_2017_CVPR}. These CNN-based methods represent a skeleton sequence as an image by encoding the temporal dynamics and the skeleton joints as rows and columns respectively, and then feed it into a CNN to recognize the underlying action just like image classification. However, in that case, only the neighboring joints within the convolutional kernel are considered to learn co-occurrence features. Although the receptive field covers all joints of a skeleton in later convolution layers, it is difficult to mine co-occurrences from all joints efficiently. Because of the weight sharing mechanism in spatial dimensions, CNN models can not learn free parameters for each joint.
% which we think preliminary information will be partially lost in the layer-wise abstraction process. 
This motivates us to design a model which is able to get global response from all joints to exploit the correlations between different joints.

We propose an end-to-end co-occurrence feature learning framework, which uses CNN to learn hierarchical co-occurrence features from skeleton sequences automatically. We find the output of a convolution layer is global response from all input channels. If each joint of a skeleton is treated as a channel, then the convolution layer can learn the co-occurrences from all joints easily.  More specifically, we represent a skeleton sequence as a tensor of shape $frames\times joints\times 3$ (the last dimension as channel). We first learn point-level features for each joint independently using convolution layers with kernel size $n\times 1$. And then we transpose the output of the convolution layers to make the dimension of $joints$ as channel. After the transpose operation, the subsequent layers aggregate global features from all joints hierarchically. Furthermore, the two-stream framework~\cite{NIPS2014_5353} is introduced to fuse the skeleton motion feature explicitly.

The main contributions of this work are summarized as follows:
\begin{itemize}
  \item We propose to employ the CNN model for learning \emph{global} co-occurrences from skeleton data, which is shown superior over local co-occurrences.
  \item We design a novel end-to-end hierarchical feature learning network, where features are aggregated gradually from point-level features to global co-occurrence features.
  \item We comprehensively exploit multi-person feature fusion strategies, which makes our network scale well to variable number of persons.
  \item The proposed framework outperforms all existing state-of-the-art methods on benchmarks for both action recognition and detection tasks.
  \end{itemize}

\section{Related Work}

The LSTM networks were designed to model long-term temporal dependency problems. Thus they are a natural choice and have been well exploited for feature learning from skeleton sequences. However, more and more literatures adopted CNN to learn skeleton features and achieved impressive performance in recent years. For example, \cite{Du2016Skeleton} proposed to cast the frame, joint and coordinate dimensions of a skeleton sequence into width, height and channel of an image respectively. Then they applied CNN for skeleton-based action recognition in the same way as CNN-based image classification. \cite{Ke_2017_CVPR} proposed an improved representation of skeleton sequences where the 3D coordinates are separated into three gray-scale images. In~\cite{Li_2017_ICMEW}, a skeleton transformer module was introduced to learn a new representation of skeleton joints. And a two-stream convolutional network was proposed to incorporate skeleton motion features.

Similar to the above works, we also adopt CNN to learn features from skeleton data. However, we attempt to model global co-occurrence patterns with CNN. And we explicitly formulate the problem into two levels of feature learning sub-problems, i.e. independent point-level feature learning and cross-joint co-occurrence feature learning.

\section{Methods}

\subsection{Co-occurrence Feature Learning with CNN}
\label{sec:global-co-occur}

\begin{figure}[tbp]
  \centering
  \includegraphics[width=0.7\linewidth]{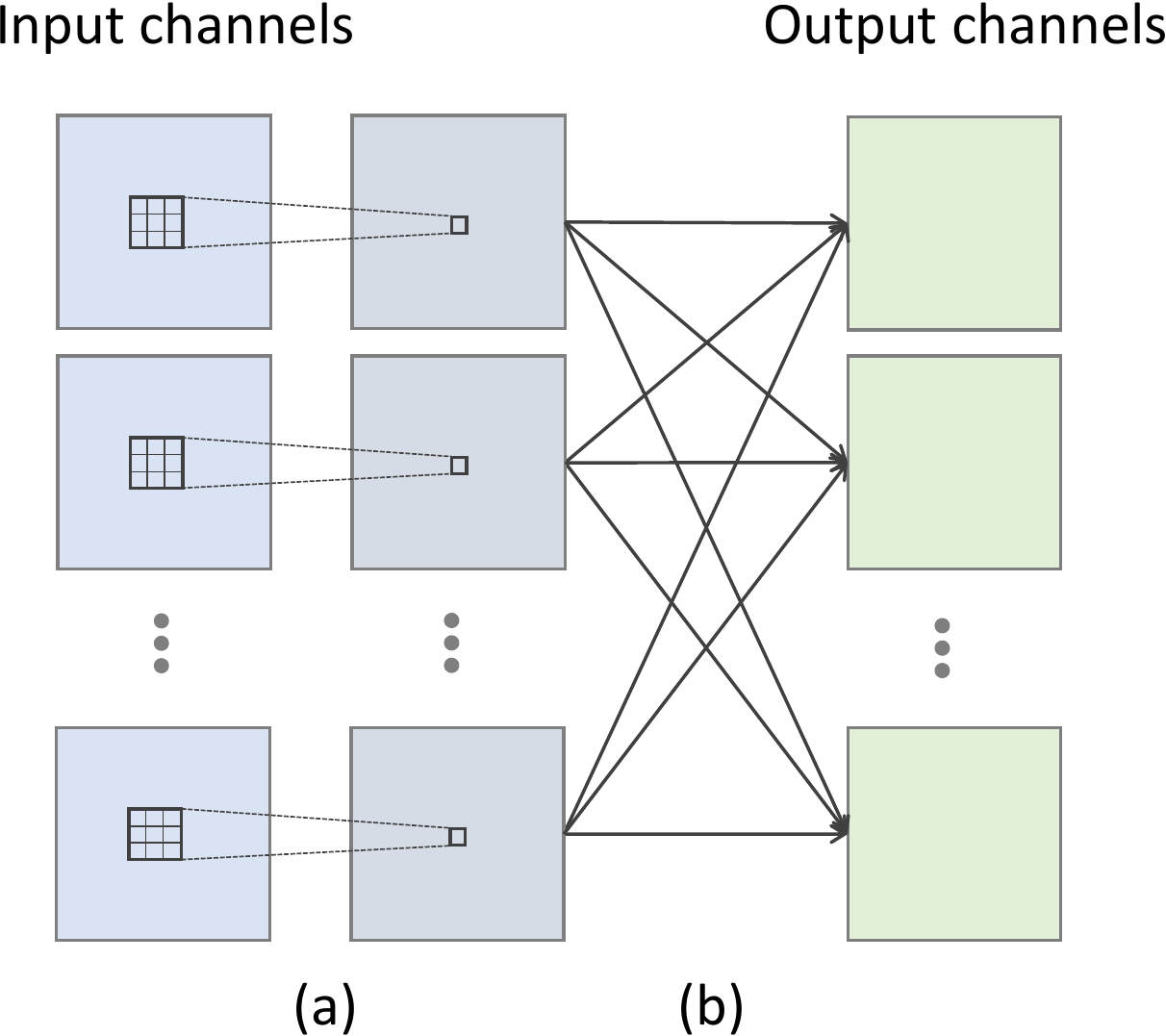}
  \caption{Decomposition of a $3\times 3$ convolution into two steps. (a) Independent 2D convolution in the spatial domain for each input channel, where features are aggregated locally from $3\times 3$ neighborhoods. (b) Element-wise summation across channels, where features are aggregated globally from all input channels.}
  \label{fig:cnn_structure}
\end{figure}

CNN is one of the most powerful and successful neural network models, which has been widely applied in image classification, object detection, video classification etc. Compared with sequential structures such as RNNs, it is capable of encoding spatial and temporal contextual information simultaneously. By investigating the convolution operation, we may decompose it into two steps (Figure~\ref{fig:cnn_structure}), i.e. local feature aggregation across the spatial domain (width and height) and global feature aggregation across channels. Then one can conclude a simple yet very practical way to regulate the aggregation extent on demand. Denote $T$ as a $d_{1}\times d_{2}\times d_{3}$ 3D tensor. We can assign different context by reorganizing (transposing) the tensor. Any information from dimension $d_{i}$ can be aggregated \emph{globally} if is specified as channels while the other two encode \emph{local} context.

In all previous CNN-based methods~\cite{Du2016Skeleton,Ke_2017_CVPR,Li_2017_ICMEW}, joint coordinates are specified as channels. It causes a problem: the co-occurrence features are aggregated locally, which may not be able to capture long-range joint interactions involved in actions like wearing a shoe. To this end, we argue that aggregating co-occurrence features \emph{globally} is of great importance and leads to better action recognition performance. It can be easily implemented by putting the joint dimension into channels of CNN's input.

\subsection{Explicit Skeleton Motion}
Besides joint co-occurrences, temporal movements of joints are crucial cues to recognize the underlying action. Although the temporal evolution pattern can be learned implicitly with CNN, we argue that an explicit modeling is preferable. Thus we introduce a representation of skeleton motion and feed it explicitly into the network.

For the skeleton of a person in frame $t$, we formulate it as $\bm{S}^t = \{\bm{J}_1^t, \bm{J}_2^t, \dots, \bm{J}_N^t\}$ where $N$ is the number of joint and $\bm{J} = (x,y,z)$ is a 3D joint coordinate. The skeleton motion is defined as the temporal difference of each joint between two consecutive frames:

\begin{align*}
  \bm{M}^t &= \bm{S}^{t+1}-\bm{S}^t \\
      &= \{\bm{J}_1^{t+1}-\bm{J}_1^t,\bm{J}_2^{t+1}-\bm{J}_2^t,...,\bm{J}_N^{t+1}-\bm{J}_N^t \}.
\end{align*}

Raw skeleton coordinates $\bm{S}$ and the skeleton motion $\bm{M}$ are fed into the network independently with a two-stream paradigm. To fuse information from the two sources, we concatenate their feature maps across channels in subsequent layers of the network (see Figure~\ref{fig:framework_HCN}).

\subsection{Hierarchical Co-occurrence Network}

\begin{figure}[tbp]
  \centering
  \includegraphics[width=0.8\linewidth]{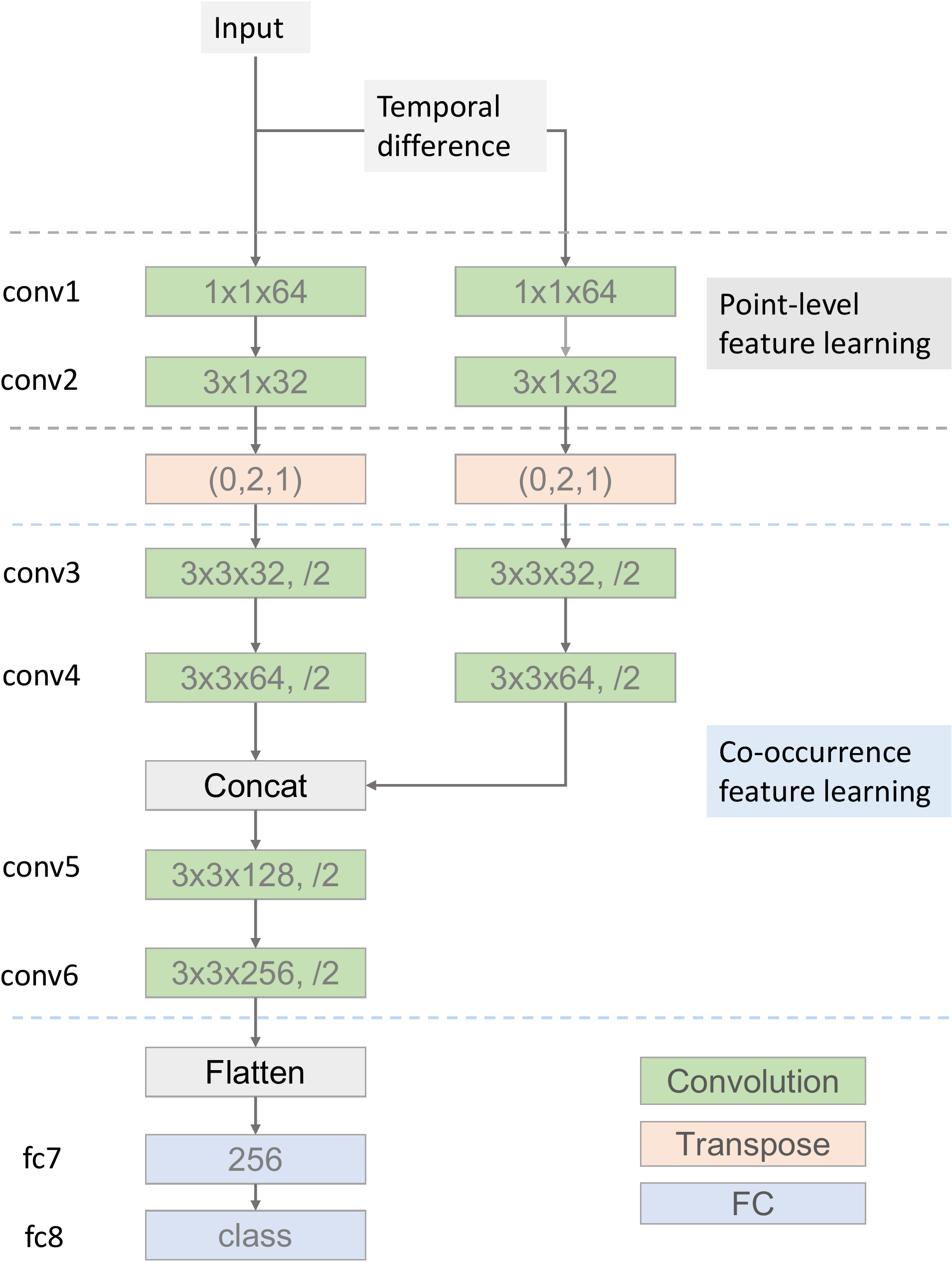}
  \caption{Overview of the proposed hierarchical co-occurrence network. Green blocks are convolution layers, where the last dimension denotes the number of output channels. A trailing ``/2'' means an appended MaxPooling layer with stride 2 after convolution. A Transpose layer permutes the dimensions of the input tensor according to the order parameter. ReLU activation function is appended after \emph{conv1}, \emph{conv5}, \emph{conv6} and \emph{fc7} to introduce non-linearity.}
  \label{fig:framework_HCN}
\end{figure}

In this section, we will elaborately describe the proposed Hierarchical Co-occurrence Network (HCN) framework, which is designed to learn the joint co-occurrences and the temporal evolutions jointly in an end-to-end manner.

Figure~\ref{fig:framework_HCN} shows the network architecture of the proposed framework. A skeleton sequence $\bm{X}$ can be represented with a $T\times N\times D$ tensor, where $T$ is the number of frames in the sequence, $N$ is the number of joints in the skeleton and $D$ is the coordinate dimension (e.g. 3 for 3D skeleton). The skeleton motion described above is of the same shape as $\bm{X}$. They are fed into the network directly as two streams of inputs. The two network branches share the same architecture. However, their parameters are not shared and learned separately. Their feature maps are fused by concatenation along channels after \emph{conv4}.

Given skeleton sequence and motion inputs, the features are learned hierarchically. In stage 1, point-level features are encoded with a $1\times 1$ (\emph{conv1}) and $n\times 1$ (\emph{conv2}) convolution layers. Since the kernel sizes along the joint dimension are kept $1$, they are forced to learn point-level representation from 3D coordinates for each joint independently. After that, we transpose the feature maps with parameter (0,~2,~1) so that the joint dimension is moved to channels of the tensor. Then in stage 2, all subsequent convolution layers extract global co-occurrence features from all joints of a person as described in Section~\ref{sec:global-co-occur}. Afterwards, the feature maps are flattened into a vector and go through two fully connected layers for final classification. Note that the skeleton's temporal evolutions are encoded all through the network with convolutions of kernel size equal to 3 along the frame dimension.

Our network contains about 0.8 million parameters, just two-thirds of the model in~\cite{Li_2017_ICMEW}. While \cite{Ke_2017_CVPR} used an ImageNet-pretrained VGG19 model. The extremely small model size allows us to easily train the network from scratch without the needs of pretraining.

\subsection{Scalability to Multiple Persons}
\label{text:multi_person}

In activities like hugging and shaking hands, multiple persons are involved. To make our framework scalable to multi-person scenarios, we perform a comprehensive evaluation on different feature fusion strategies.

{\bfseries Early fusion.} All joints from multiple persons are stacked as input of the network. For variable number of persons, zero padding is applied if the number of persons is less than the pre-defined maximal number.

\begin{figure}[tbp]
  \centering
  \includegraphics[width=\linewidth]{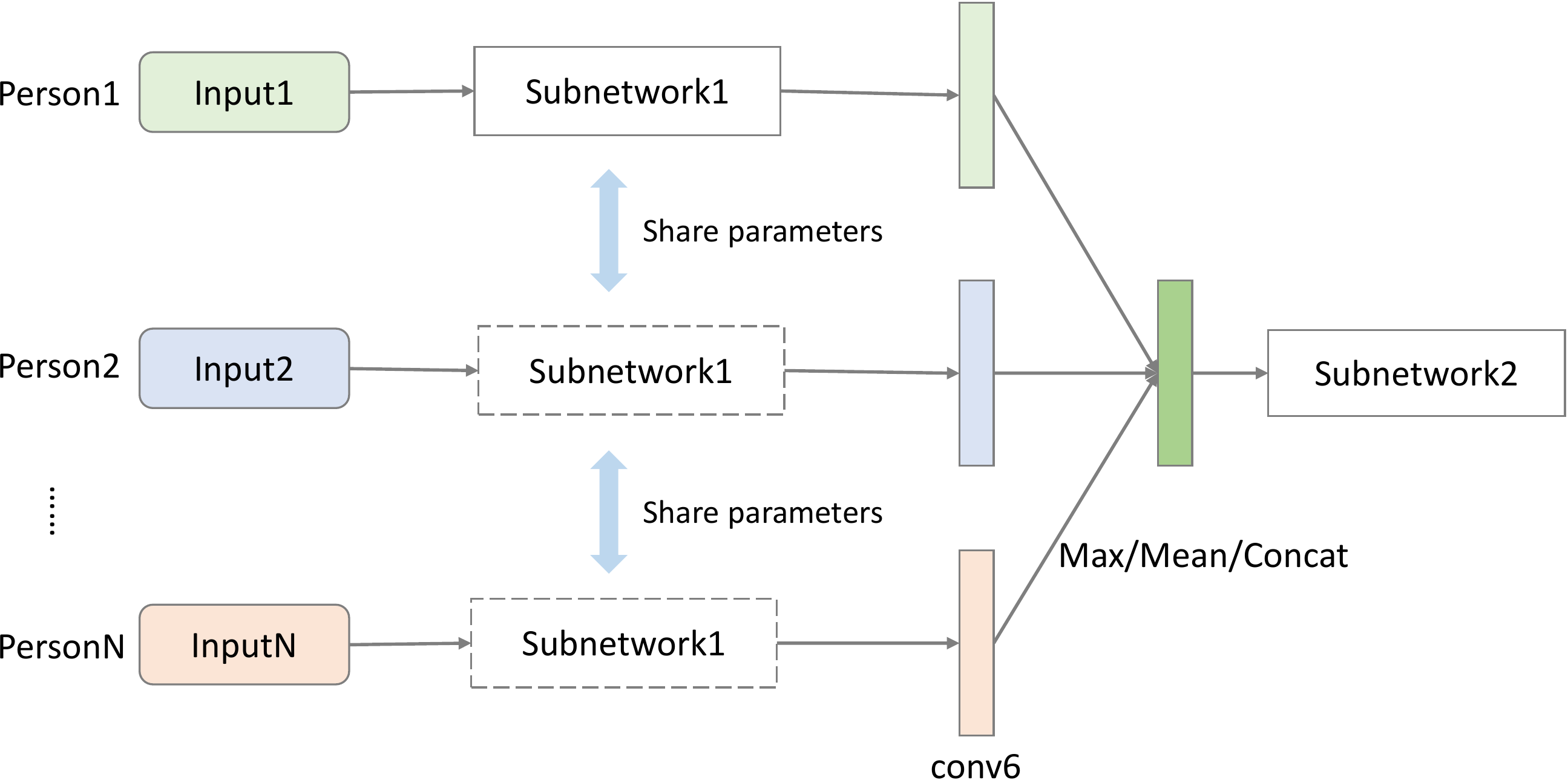}
  \caption{Late fusion diagram for multi-person feature fusion. Maximum, mean and concatenation operations are evaluated in terms of performance and generalization.
  \label{fig:multi_person}}
\end{figure}

{\bfseries Late fusion.} As illustrated in Figure~\ref{fig:multi_person}, inputs of multiple persons go through the same subnetwork and their \emph{conv6} feature maps are merged with either concatenation along channels or element-wise maximum / mean operation.

Note that element-wise late fusion can generalize well to variable number of persons, while the others require a pre-defined maximal number. Besides, compared with single person, no extra parameter is introduced.

\subsection{Action Recognition and Detection}

\begin{figure*}[htbp]
  \centering
  \includegraphics[width=0.8\linewidth]{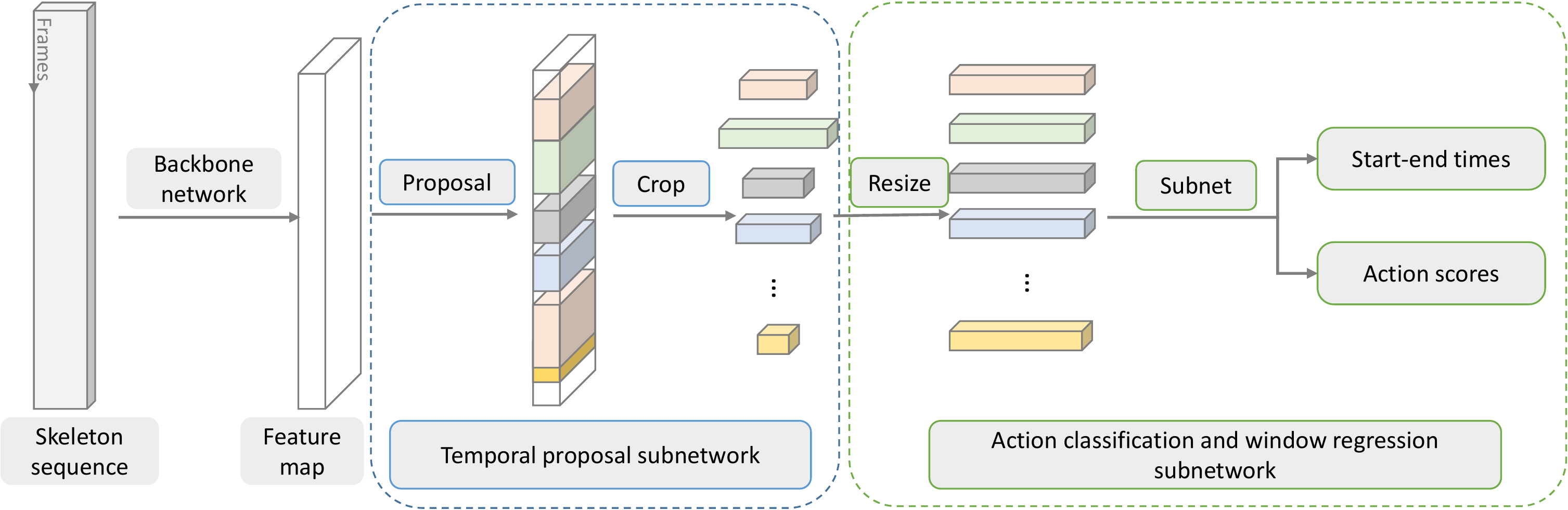}
  \caption{The temporal action detection framework. The backbone network is described in Figure~\ref{fig:framework_HCN}. Two subnetworks are designed for temporal proposal segmentation and action classification respectively.
  \label{fig:DET}}
\end{figure*}

For the recognition task, a softmax function is used to normalize the output of the network. The probability that a sequence $\bm{X}$ belongs to the $i^{th}$ class is
\begin{align}
  P(C_i|\bm{X})=\frac{e^{o_i}}{\sum_{j=1}^{C}e^{o_j}}, \quad i=1, 2, \dots, C,
\end{align}
where $\bm{o}=(o_1, o_2, \dots, o_c)^T$ is the output of the network, $C$ is the number of classes.

We also extend the proposed network for temporal action detection. Previously the Faster R-CNN framework~\cite{Ren2015Faster} has been adapted for the task of temporal action detection~\cite{Rc3d,Li_2017_ICMEW}. Following these works, we briefly introduce our implementation and later show that with the proposed hierarchical co-occurrence features, detection performance also gets significantly improved. The detection framework is shown in Figure~\ref{fig:DET}. Specifically, based on the backbone feature learning network in Figure~\ref{fig:framework_HCN}, two subnetworks are appended after \emph{conv5}, i.e. the temporal proposal subnetwork and the action classification subnetwork. The temporal proposal subnetwork predicts variable-length temporal segments that potentially contain an action. The corresponding feature maps of each proposal are extracted using a crop-and-resize operation. After that, the classification subnetwork predicts their action categories. In both subnetworks, window regression is performed to obtain more accurate localization. We use softmax loss for classification and smooth $L_1$ loss for window regression. The two tasks are optimized jointly. The objective function is given by:
\begin{align}
  L=\frac{1}{N_{cls}}\sum_{i}L_{cls}(p_i,p_i^*)+
  \lambda \frac{1}{N_{reg}}\sum_{i}L_{reg}(t_i,t_i^*).
  \label{eq:reg_loss}
\end{align}
$N_{cls}$ and $N_{reg}$ are the number of samples. $\lambda$ is the weight applied on the regression loss (empirically set to 1 in our experiments). $i$ is the index of an anchor or proposal in a batch. $p$ is the predicted probability and $p^*$ is the groundtruth label. $t=\{t_x, t_w\}$ is the predicted regression target and $t^*=\{t_x^*, t_w^*\}$ is the groundtruth target. Window regression is essentially an adaptation of bounding box regression~\cite{girshick2014rich}, where targets along one dimension rather than two are predicted. The target transformations are computed as follows:
\begin{align}
  t_x=(x-x_a)/w_a,~t_w=\log(w/w_a),\\
  t_x^*=(x^*-x_a)/w_a,~t_w^*=\log(w^*/w_a),
\end{align}
where $x$ and $w$ denote the center and length of the temporal window. And $x$, $x_a$ and $x^*$ represent the predicted window, anchor window and groundtruth window respectively. The same rule applies for $w$.

\section{Experiments}
We evaluate the proposed method on three common benchmark datasets, i.e. the NTU RGB+D~\cite{NTURGBD} and SBU Kinect Interaction~\cite{SBU} datasets for action recognition, and the PKU-MMD~\cite{PKUMMD} dataset for temporal action detection. Besides, an ablation study is performed to show the importance of global co-occurrence feature aggregation.

\subsection{Datasets and Implementation Details}
\subsubsection{NTU RGB+D}

The NTU RGB+D dataset is so far the largest skeleton-based human action recognition dataset. It contains 56880 skeleton sequences, which are annotated as one of 60 action classes. There are two recommended evaluation protocols, i.e. Cross-Subject (CS) and Cross-View (CV). In the cross-subject setting, sequences of 20 subjects are used for training and sequences of the rest 20 subjects are used for validation. In the cross-view setting, samples are split by camera views. Samples from two camera views are used for training and the rest are used for testing.

During training, we randomly crop a sub-sequence from the entire sequence. The cropping ratio is drawn from uniform distribution between [0.5, 1]. During inference, we center crop a sub-sequence with a ratio of 0.9. Since different actions last for various durations, the input sequences are normalized to a fixed length (32 in our experiments) with bilinear interpolation along the frame dimension. To alleviate the problem of overfitting, we append dropout after \emph{conv4}, \emph{conv5}, \emph{conv6} and \emph{fc7} with a dropout ratio of 0.5. A weight decay of 0.001 is applied on the weights of \emph{fc7}. We train the model for 300k iterations in total, with a mini-batch size of 64. The Adam~\cite{kingma2015adam} optimizer is utilized. The learning rate is initialized to 0.001 and exponentially decayed every 1k steps with a rate of 0.99.

\subsubsection{SBU Kinect Interaction}
The SBU Kinect Interaction dataset~\cite{SBU} is a Kinect captured human activity recognition dataset depicting two person interaction. It contains 282 skeleton sequences and 6822 frames of 8 classes. There are 15 joints for each skeleton. For evaluation we perform subject-independent 5-fold cross validation as suggested in~\cite{SBU}.

Considering the small size of the dataset, we simplify the network architecture in Figure~\ref{fig:framework_HCN} accordingly. Specifically, the output channels of \emph{conv1}, \emph{conv2}, \emph{conv3}, \emph{conv5}, \emph{conv6} and \emph{fc7} are reduced to 32, 16, 16, 32, 64 and 64 respectively. And the \emph{conv4} layer is removed. Besides, all the input sequences are normalized to a length of 16 frames rather than 32.

\subsubsection{PKU-MMD}
The PKU-MMD dataset~\cite{PKUMMD} is currently the largest skeleton-based action detection dataset. It contains 1076 long untrimmed video sequences performed by 66 subjects in three camera views. 51 action categories are annotated, resulting almost 20,000 action instances and 5.4 million frames in total. Similar to NTU RGB+D, there are also two recommended evaluate protocols, i.e. cross-subject and cross-view. For detection-specific hyper-parameters, we basically follow the settings in~\cite{Ren2015Faster}. In particular, we use anchor scales of \{50, 100, 200, 400\} in the temporal proposal network.

\subsection{Multi-person Feature Fusion}
To evaluate different ways of multi-person feature fusion described in Section~\ref{text:multi_person}, we perform an ablation study on the NTU RGB+D dataset. As shown in Table~\ref{table:multi_person_fusion}, all late fusion methods outperform the early fusion method. The reason might be that features of different persons are better aligned and thus more compatible in high-level semantic space than in raw input space. Among the three late fusion implementations, the element-wise maximum operation achieves the best accuracy. This is due to the side effect of zero padding for single-person actions. That is, compared with multi-person samples, features of single-person samples get weakened with the padded zeros in both cases of concatenation and element-wise mean. While element-wise maximum does not suffer from this issue. In the following experiments the late fusion with element-wise maximum strategy is adopted.

\begin{table}[htbp]
  \centering
  \begin{tabular}{c|l|c}\hline
    \multicolumn{2}{c|}{Method} & Accuracy (\%) \\ \hline
    \multicolumn{2}{c|}{Early fusion} & 85.2 \\ \hline
    \multirow{3}{*}{Late fusion} &Mean &85.8 \\ 
     &Concat &85.9 \\ 
    &{\bfseries Max} &{\bfseries 86.5} \\ \hline
  \end{tabular}
  %\vspace{-2mm}
  \caption{Performance of different fusion methods for multi-person feature on the NTU RGB+D dataset in the cross-subject setting.}
  \label{table:multi_person_fusion}
\end{table}

\subsection{Comparison to Other State-of-the-arts}
A systematic evaluation of the proposed HCN framework is performed on the three datasets mentioned above. As shown in Table~\ref{table:NTU}, Table~\ref{table:SBU} and Table~\ref{table:PKU}, our approach consistently outperforms the current state-of-the-arts in terms of both action recognition accuracy and action detection mAP.

On the large-scale NTU RGB+D dataset (Table~\ref{table:NTU}), our approach achieves the best action recognition accuracy. Compared with the state-of-the-art LSTM-based method~\cite{view_adaptive}, the accuracy is improved by 7.3\% in the cross-subject setting and 3.4\% in the cross-view setting. Compared with the most recent two-stream CNN method~\cite{Li_2017_ICMEW}, the accuracy is improved by 3.3\% in the cross-subject setting and 1.8\% in the cross-view setting. See Figure~\ref{fig:cls_examples} for visualization of our exemplary classifications.

\begin{table}[tbp]
  \centering
  \begin{tabular}{c|cc}\hline
    \multirow{2}{*}{Methods} & \multicolumn{2}{c}{Accuracy (\%)} \\
    & CS & CV \\ \hline
    Deep LSTM \cite{NTURGBD} &60.7 &67.3 \\
    Part-aware LSTM \cite{NTURGBD}     &62.9 &70.3 \\
    ST-LSTM+Trust Gate \cite{trust_gate} &69.2 &77.7 \\
    STA-LSTM \cite{song2016end}   &73.4 &81.2 \\
    Clips + CNN + MTLN \cite{Ke_2017_CVPR}  &79.6 &84.8 \\
    VA-LSTM \cite{view_adaptive}    &79.2  &87.7 \\
    Two-stream CNN \cite{Li_2017_ICMEW} &83.2  &89.3 \\
    Proposed {\bfseries HCN}  &{\bfseries 86.5} &{\bfseries 91.1} \\
    \hline
  \end{tabular}
  %\vspace{-2mm}
  \caption{Action classification performance on the NTU RGB+D dataset. CS and CV mean the cross-subject and cross-view settings respectively.}
  \label{table:NTU}
\end{table}

On the small SBU Kinect Interaction dataset (Table~\ref{table:SBU}), the proposed method also outperforms other methods by a large margin. Compared with the LSTM-based co-occurrence learning baseline~\cite{co_occurrence}, the accuracy is improved by 8.2\%, which proves the superiority of our CNN-based global co-occurrence feature learning framework. Since the recognition task on SBU is much easier than on NTU RGB+D, the best accuracy reported has achieved 97.6\%~\cite{view_adaptive}. Yet we push the frontier further and achieve an accuracy of 98.6\%.

\begin{table}[tbp]
  \centering
  \begin{tabular}{c|c}\hline
    Methods & Accuracy (\%) \\ \hline
    Raw skeleton \cite{Ji2014Interactive} &79.4 \\
    Joint feature \cite{Ji2014Interactive} &86.9 \\
    ST-LSTM \cite{trust_gate} &88.6 \\
    Co-occurrence RNN \cite{co_occurrence} &90.4 \\
    STA-LSTM \cite{song2016end} &91.5 \\
    ST-LSTM+Trust Gate \cite{trust_gate} &93.3 \\
    VA-LSTM \cite{view_adaptive} &97.6 \\
    Proposed \textbf{HCN}  & \textbf{98.6} \\
    \hline
  \end{tabular}
  %\vspace{-2mm}
  \caption{Action classification performance on the SBU dataset.}
  \label{table:SBU}
\end{table}

On the PKU-MMD action detection dataset (Table~\ref{table:PKU}), we also achieve state-of-the-art performance. Compared with the Skeleton boxes method~\cite{libo}, in which an adapted VGGNet is employed for temporal action detection, our method improve the mAP by 38\% in the cross-subject setting. Compared with the recent work in~\cite{Li_2017_ICMEW}, where a similar detection framework is utilized, our method improves the mAP by 2.2\% and 0.5\% in the cross-subject and cross-view settings respectively. Note that our improvement over it is purely owing to better features learned with the proposed framework. See Figure~\ref{fig:det_examples} for visualization of our exemplary detections.

\begin{table}[tbp]
  \centering
  \begin{tabular}{c|cc}\hline
    \multirow{2}{*}{Methods} & \multicolumn{2}{c}{mAP (\%)} \\
    & CS & CV \\ \hline
    STA-LSTM \cite{song2016end} &44.4 &13.1 \\
    JCRRNN \cite{li2016online} &32.5 &53.3 \\
    Skeleton boxes \cite{libo}  &54.8 &94.2 \\
    Li et al. \cite{Li_2017_ICMEW} &90.4 &93.7 \\
    Proposed \textbf{HCN}  &\textbf{92.6} &\textbf{94.2} \\
    \hline
  \end{tabular}
  %\vspace{-2mm}
  \caption{Action detection performance on the PKU-MMD dataset. mAP is measured at an IoU threshold of 0.5.}
  \label{table:PKU}
\end{table}

From the results, we can conclude that our proposed hierarchical co-occurrence feature learning framework is superior over other state-of-the-art LSTM and CNN-based methods. It is scalable to datasets of different sizes. And the learned features generalize well across various tasks, from action classification to action detection.

\subsection{Importance of Global Co-occurrence Feature Learning}

To further understand the behavior of global co-occurrence feature learning, we perform an ablation study. Specifically, we deliberately prepend a Transpose layer with parameter (0,~2,~1) before \emph{conv1} so that the joint and coordinate dimensions are swapped. Then in subsequent layers after \emph{conv2}, co-occurrence features are aggregated locally rather than globally, as in previous CNN-based methods~\cite{Du2016Skeleton,Ke_2017_CVPR,Li_2017_ICMEW}. The modified network is referred to as HCN-local. We train HCN-local models with the same hyper-parameters as HCN. The results are listed in Table~\ref{table:HPN_HCN}. We can see that HCN consistently outperforms HCN-local on all of the three datasets. In particular, performance gain of HCN over HCN-local is more significant in the case of cross-subject than cross-view. This observation indicates that with global co-occurrence feature, the variation across different persons can be well addressed.

\begin{table}[htbp]
  \centering
  \begin{tabular}{c|cc|c|cc}\hline
    \multirow{2}{*}{Methods} & \multicolumn{2}{c|}{NTU RGB+D} & SBU & \multicolumn{2}{c}{PKU-MMD} \\
              & CS   & CV   & -    & CS   & CV   \\ \hline
    HCN-local & 83.9 & 89.7 & 96.8 & 91.1 & 93.9 \\
    HCN       & 86.5 & 91.1 & 98.6 & 92.6 & 94.2 \\
    \hline
  \end{tabular}
  %\vspace{-2mm}
  \caption{Comparison of HCN-local and HCN in terms of classification accuracy on the NTU RGB+D and SBU datasets and detection mAP on the PKU-MMD dataset.}
  \label{table:HPN_HCN}
\end{table}

\begin{figure}[htbp]
  \centering
  \includegraphics[width=\linewidth]{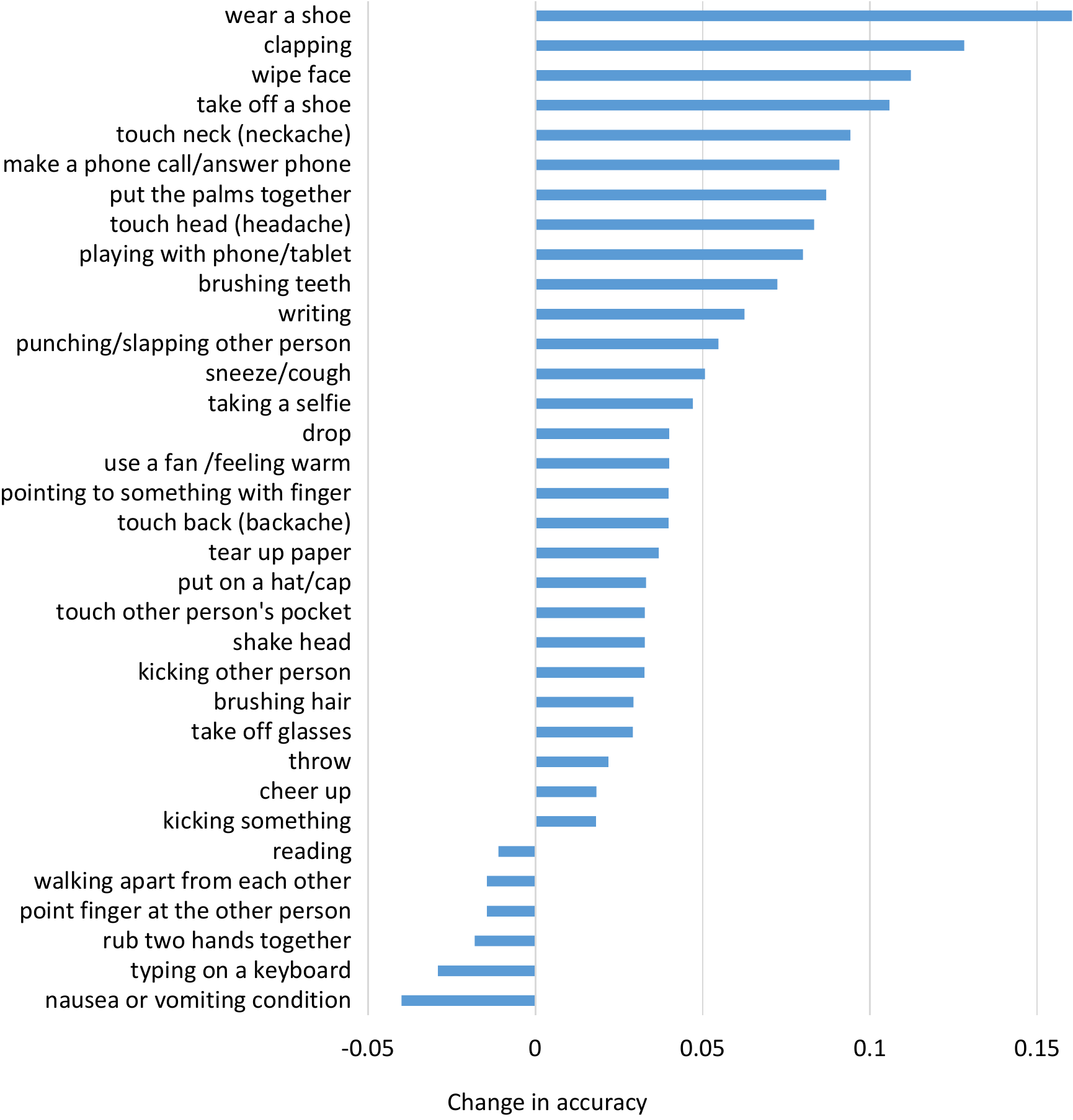}
  \caption{Per-category change in accuracy of HCN over HCN-local on the NTU RGB+D dataset in the cross-subject setting. For clarity only categories with change greater than 1\% are shown.}
  \label{fig:HCN-HPN-by-class}
\end{figure}

For a detailed comparison, we further investigate the per-category change in accuracy. Figure~\ref{fig:HCN-HPN-by-class} shows the results, where the categories are sorted by accuracy gain. We can see that most actions get improved, in particular those involving long-range joint interaction. For example, over 10\% absolute improvement is observed for \emph{wear a shoe}, \emph{clapping}, \emph{wipe face} and \emph{take-off a shoe}. For actions with no obvious joint interaction such as \emph{nausea} and \emph{typing on a keyboard}, global co-occurrence feature is not critical.

\section{Discussion}
What motivates us to explore the different levels of context aggregation is to figure out the importance of modeling interactions among joint points in action recognition. Given a certain type of action of one or more subjects, do the whole or partial interactions of joints contribute to the recognition on earth? The answer from our experiments is initially counter-intuitive but makes sense, which has been proven by many analogous works~\cite{he2016deep,zhong2017cascade}. The so-called background context is an essential factor for boosting a task's performance and it is the same case in action recognition. For recognition of a specific action, e.g. make a phone call, the joints of uninterest, say ankle, play a similar role as background context and the contribution is encoded implicitly with CNNs. This is just the insights which our method benefits from.

\section{Conclusions}
We present an end-to-end hierarchical co-occurrence feature learning framework for skeleton-based action recognition and detection. By exploiting the capability to global aggregation of CNN, we find that the joint co-occurrence features can be learned with CNN model simply and efficiently. In our method, we learn point-level features for each joint independently, Afterwards, we treat the  feature of each joint as a channel of convolution layer to learn hierarchical co-occurrence features. And a two-stream framework is adopted to fuse the motion feature. Furthermore, we exploit the best way to deal with multi-person involved activities. Experiments on three benchmark datasets demonstrate that the proposed HCN model significantly improves the performance on both action recognition and detection tasks.

\bibliographystyle{named}
\bibliography{ijcai18_ref}

\newpage
\onecolumn
\appendix

\section{Exemplary Results}

\begin{figure*}[h!]
  \centering
  \includegraphics[width=0.9\linewidth]{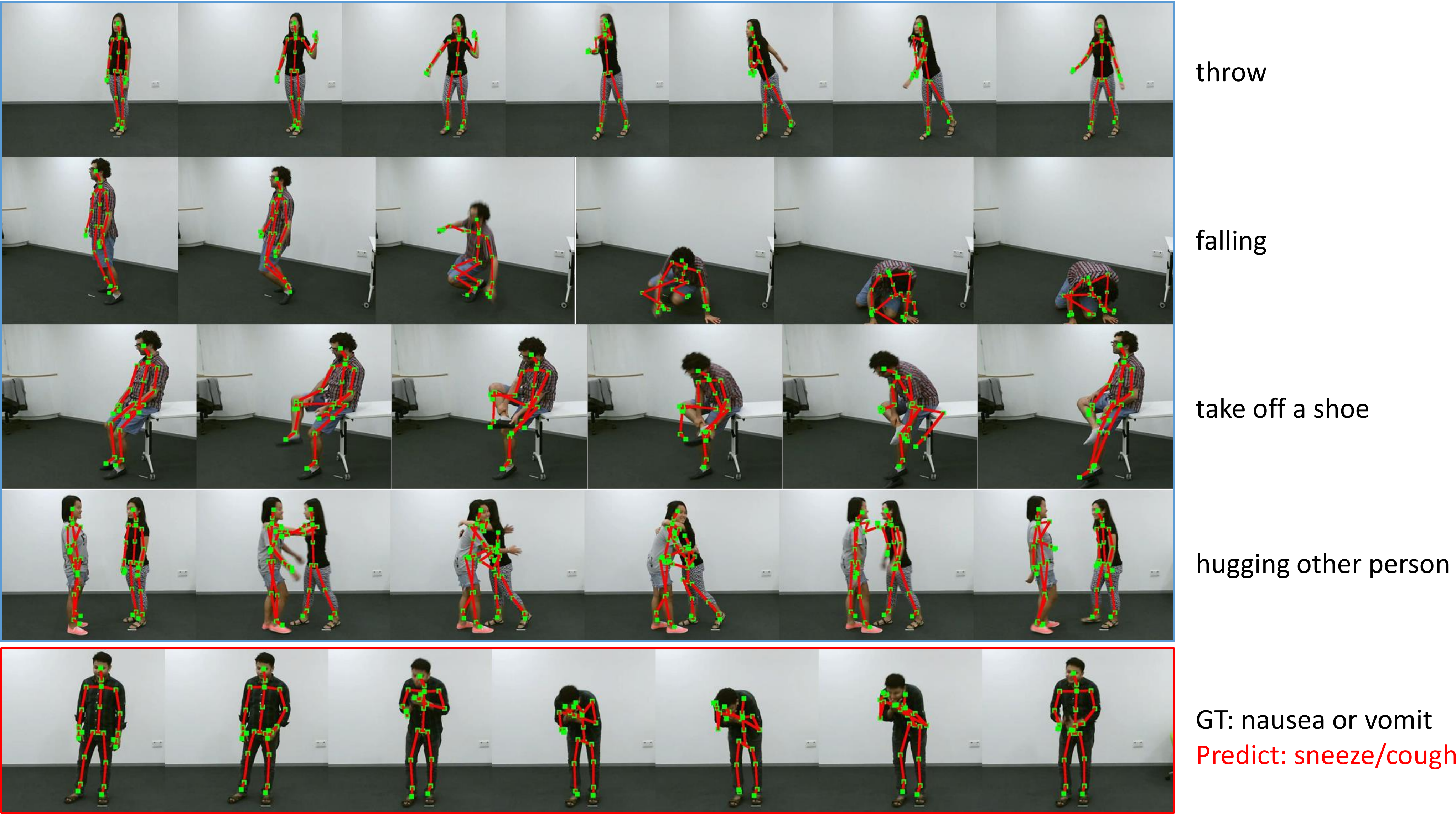}
  \caption{Exemplary action recognition results on the val set of the NTU RGB+D dataset. The upper four samples are correctly recognized. And a failure case is shown in the bottom row.}
  \label{fig:cls_examples}
\end{figure*}

\begin{figure*}[h!]
  \centering
  \includegraphics[width=\linewidth]{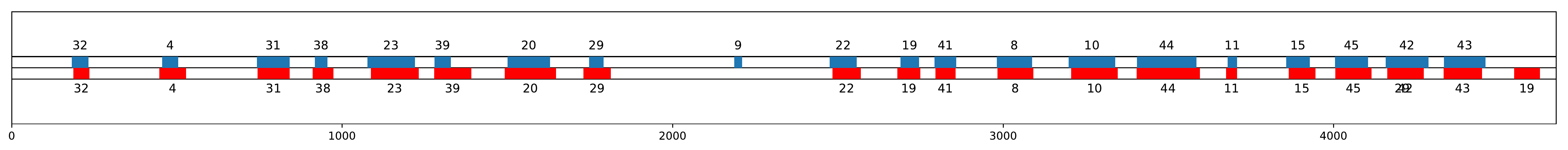} \\
  \includegraphics[width=\linewidth]{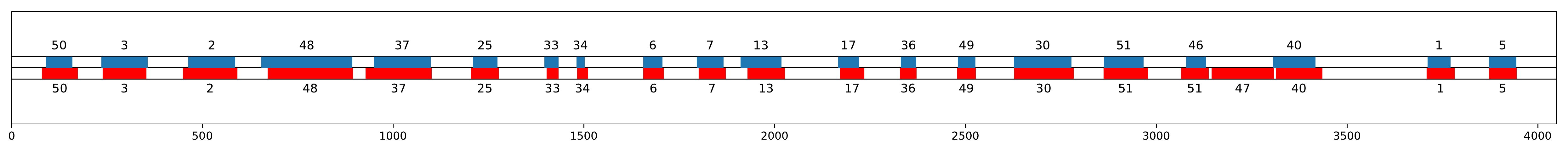} \\
  \includegraphics[width=\linewidth]{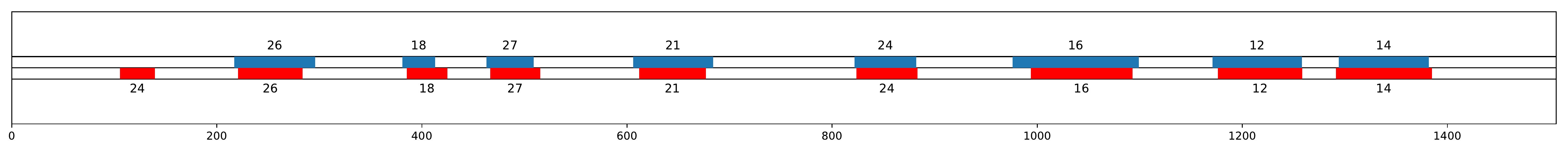} \\
  \includegraphics[width=\linewidth]{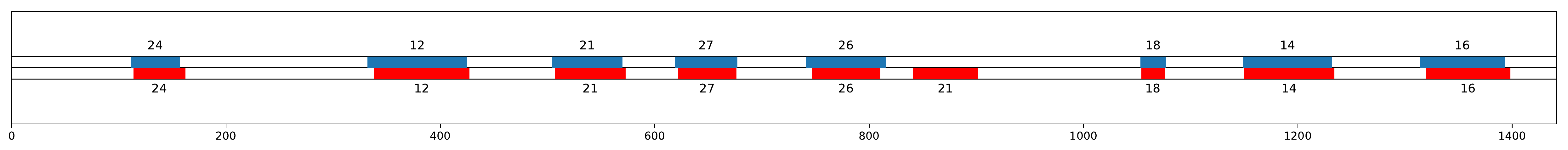} \\
  \includegraphics[width=\linewidth]{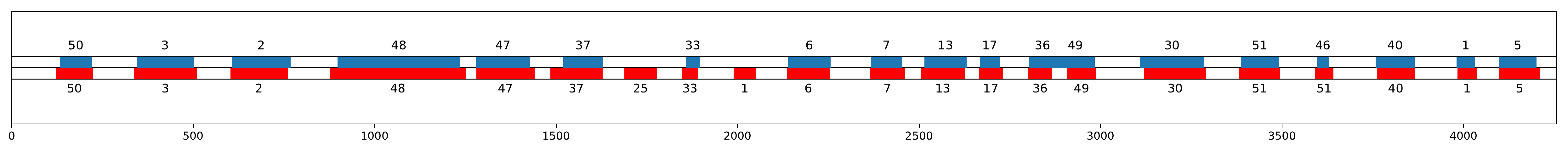} \\
  \caption{Exemplary action detection results on the val set of the PKU-MMD dataset. Five samples are shown. For each sample, the horizontal axis represents frame indices. Groundtruth action segments are drawn in blue, while detected segments with confidence greater than 0.6 are displayed in red. The number attached to each segment is the category ID. Best viewed in color.}
  \label{fig:det_examples}
\end{figure*}

\end{document}